\documentclass[twocolumn,10pt]{article}

\usepackage[margin=0.9in,columnsep=0.28in]{geometry}
\usepackage[T1]{fontenc}
\usepackage{mathptmx}
\usepackage{amsmath,amssymb}
\usepackage{booktabs}
\usepackage{multirow}
\usepackage{graphicx}
\usepackage{xcolor}
\usepackage{microtype}
\usepackage{tikz}
\usetikzlibrary{positioning,arrows.meta,shapes.misc}
\usepackage{pgfplots}
\pgfplotsset{compat=1.18}
\usepackage[colorlinks=true,linkcolor=blue!60!black,citecolor=blue!60!black,urlcolor=blue!60!black]{hyperref}
\usepackage{caption}
\usepackage[numbers,sort&compress]{natbib}

\definecolor{gitgreen}{RGB}{47,129,90}
\definecolor{nullred}{RGB}{178,60,60}
\definecolor{boxgray}{RGB}{245,245,245}

\newcommand{\pp}{\,\mathrm{pp}}
\newcommand{\ci}[2]{[#1,\,#2]}
\newcommand{\sig}{$^{*}$}

\title{\textbf{GitOfThoughts: Version-Controlled Reasoning and Agent Memory\\ You Can Replay, Diff, and Merge}}

\author{
Pavan C Shekar \quad Abhishek H S \quad Aswanth Krishnan \\[3pt]
\normalsize QpiAI, Bengaluru, India \\[1pt]
\normalsize \texttt{\{pavan.s,\,abhishek.hs,\,ashwanth.krishnan\}@qpiai.tech}
}

\date{}

\begin{document}

\maketitle

\begin{abstract}
Large language model reasoning leaves no trace once it is done. The steps of a chain of thought disappear when the context window closes, a pruned search branch is just gone, and memory buffers cannot be diffed, merged, or audited. Code, infrastructure, and experiments are all version-controlled. Reasoning is not. GitOfThoughts stores an agent's reasoning tree as a git repository. Every scored thought becomes a commit, scores become notes, outcomes become tags, and retrieval is just \texttt{git log} over the agent's own history.

We use this to test something simple. Does giving an agent memory from past problems actually make it more accurate? We tried five memory stores (none, a markdown file, a vector database, a graph, and git) across two benchmarks, two model sizes, and several pre-registered repeat experiments. The answer, on new problems, is no, including one promising early result that did not hold up when we repeated it. Memory only helps once the problem being solved is nearly identical to something already in memory (cosine similarity above about 0.8); below that, it does nothing. In other words, the model is finding the answer rather than learning the method. Even a model $4.5\times$ larger still cannot pull a reusable method out of a worked example; it just gets better at spotting near-copies. The only thing that reliably helped on new problems was generating several answers and picking the most common one (self-consistency). So the case for using git as the memory store is not that it retrieves better. It is that it gives auditability, history, and the ability to merge two agents' memories, at no cost to accuracy.
\end{abstract}

\section{Introduction}
\label{sec:intro}

Modern software engineering rests on a substrate so universal it has become invisible: every change to a non-trivial codebase is an immutable, parent-linked, content-addressed object in a version-control system. The same now holds for infrastructure-as-code, scientific datasets, and experiment tracking \citep{zaharia2018mlflow}. \emph{Reasoning} is the remaining outlier. When a chain-of-thought prompt \citep{wei2022cot} expires, its steps are gone; when a tree-of-thoughts agent \citep{yao2023tot} prunes a branch, the pruned reasoning leaves no record; when Reflexion \citep{shinn2023reflexion} writes a self-criticism, that buffer has no diff, no merge, no signed author, and no way for a third party to verify what the agent thought happened.

This gets in the way of real work. It blocks \emph{reproducibility} (replaying ``what did the agent think at step 17?''), \emph{audit} (detecting train--test leakage or gold-answer memorization), \emph{memory transfer} (combining two agents' experience), and \emph{incident review} (retracing a confident wrong answer means re-running it, not reading its history).

This paper tests two falsifiable hypotheses. \textbf{H-substrate:} a version-controlled reasoning substrate provides operational value (replay, audit, diff, merge) at accuracy parity with conventional memory stores. \textbf{H-memory:} cross-problem memory improves an agent's accuracy on novel problems. GitOfThoughts is the instrument for both: the reasoning tree is a git repository (\S\ref{sec:design}), and a pluggable memory interface lets us hold the agent fixed while swapping only the substrate. The bulk of the paper is a controlled, pre-registered evaluation of these hypotheses. Our main findings are:

\begin{enumerate}
\item \textbf{H-memory is rejected on novel problems.} Across two benchmarks (GPQA-Diamond, MATH-500), two transfer regimes (cross-problem, cross-episode), two backbones, and up to $n{=}500$, \emph{no} memory substrate reliably improves accuracy on novel problems (\S\ref{sec:null}). A promising $+15\pp$ git trend at $n{=}40$ did not survive its own pre-registered replication.
\item \textbf{Memory works in exactly one regime: near-duplicate retrieval.} A similarity sweep shows memory helps sharply ($+12$ to $+13.5\pp$\sig) once the retrieved case is a near-duplicate of the test problem (cosine $\gtrsim 0.8$), and nothing below that. The gain is answer retrieval, not method transfer. A $4.5\times$ stronger backbone does not change this: scale steepens the copyability step (near-duplicate gain $+22.5$ to $+28.5\pp$\sig) while the method band stays null (\S\ref{sec:copyability}). This single result explains every null in the paper and bounds when agent memory pays off: recurring workloads, not novel problems.
\item \textbf{Test-time sampling, not memory, is what moves accuracy.} Self-consistency gives $+3.4\pp$\sig{} at $n{=}500$. Our own system headline on GPQA (47.0\% vs.\ 33.0\%) looks better than that, but the gain comes from MCQ-aware expansion plus a much larger compute budget (\S\ref{sec:levers}), not from git or memory, and we say so rather than claim it as a win.
\item \textbf{H-substrate is supported.} Git delivers auditability, provenance, line-level diffs over reasoning text, deterministic replay, and mergeable memory, at accuracy parity with every other substrate and at small absolute cost ($\sim$15\,ms/write; \S\ref{sec:design}, \S\ref{sec:audit}).
\end{enumerate}

We document a measurement bug we caught, a result we retracted, and a hypothesis the data refuted, with the git history behind each.

\paragraph{What GitOfThoughts is not.} We claim no new tree-search algorithm; the outer tree-of-thoughts and inner ReAct \citep{yao2023react} loops are minimum-sufficient plumbing. We do not claim git is the only valid substrate, only that it is the most mature, operationally rich, and widely deployed content-addressed store available. The thesis: \emph{reasoning is the last unversioned software process; GitOfThoughts runs \texttt{git init} on it.}

\section{Related Work}
\label{sec:related}

\textbf{Iterative reasoning.} ReAct \citep{yao2023react}, Reflexion \citep{shinn2023reflexion}, Self-Refine \citep{madaan2023selfrefine}, and self-consistency \citep{wang2023sc} enrich single-question traces but keep them in transient in-memory structures that vanish at episode end and expose no transfer operation.

\textbf{Tree search.} Tree-of-Thoughts \citep{yao2023tot}, LATS \citep{zhou2024lats}, RAP \citep{hao2023rap}, and ReTreVal \citep{retreval2026} extend search with reflection, MCTS, planning, or validation and cross-problem memory, but in each the tree is a Python object freed at process exit; GitOfThoughts gives this family a persistent, versioned substrate.

\textbf{Long-term memory.} Voyager's skill library \citep{wang2023voyager}, Generative Agents' observation stream \citep{park2023generative}, MemoryBank \citep{zhong2024memorybank}, MemGPT \citep{packer2023memgpt}, ExpeL \citep{zhao2024expel}, CLIN \citep{majumder2023clin}, Mem0 \citep{chhikara2025mem0}, A-MEM \citep{xu2025amem}, and TextGrad \citep{yuksekgonul2024textgrad} all use bespoke side stores. We map this landscape onto our \{markdown, vector, graph\} arms and add git as a versioned substrate. GitOfThoughts differs in substrate: its memory \emph{is} the reasoning DAG itself, in a content-addressed VCS. To our knowledge, no prior work treats the commit log of an LLM agent as its runtime memory, or empirically compares a VCS against vector/graph stores as a reasoning-memory substrate.

\textbf{Test-time scaling.} Self-consistency \citep{wang2023sc}, multi-agent debate \citep{du2023debate}, and verifier-based best-of-$N$ selection \citep{snell2024testtime} trade compute for accuracy. Recent work finds debate mainly improves individual predictions while adjudication is the hard part, consistent with our selector-is-the-bottleneck result (\S\ref{sec:levers}).

\textbf{In-context learning mechanisms.} Our copyability finding connects to evidence that demonstrations often act through task recognition and surface format rather than method content \citep{min2022rethinking}. In our controlled arms, the content of a retrieved worked example barely matters unless the example is copyable, which is the memory-substrate analogue of that ICL result.

\textbf{Benchmarks.} GPQA-Diamond \citep{rein2023gpqa} (graduate-level, ``Google-proof'' multiple choice), MATH-500 \citep{hendrycks2021math}, and ScienceWorld \citep{wang2022scienceworld}, where SwiftSage \citep{lin2023swiftsage} and CLIN report cross-episode learning.

\section{Design: Reasoning as a Versioned DAG}
\label{sec:design}

\begin{figure*}[t]
\centering
\includegraphics[width=0.92\textwidth]{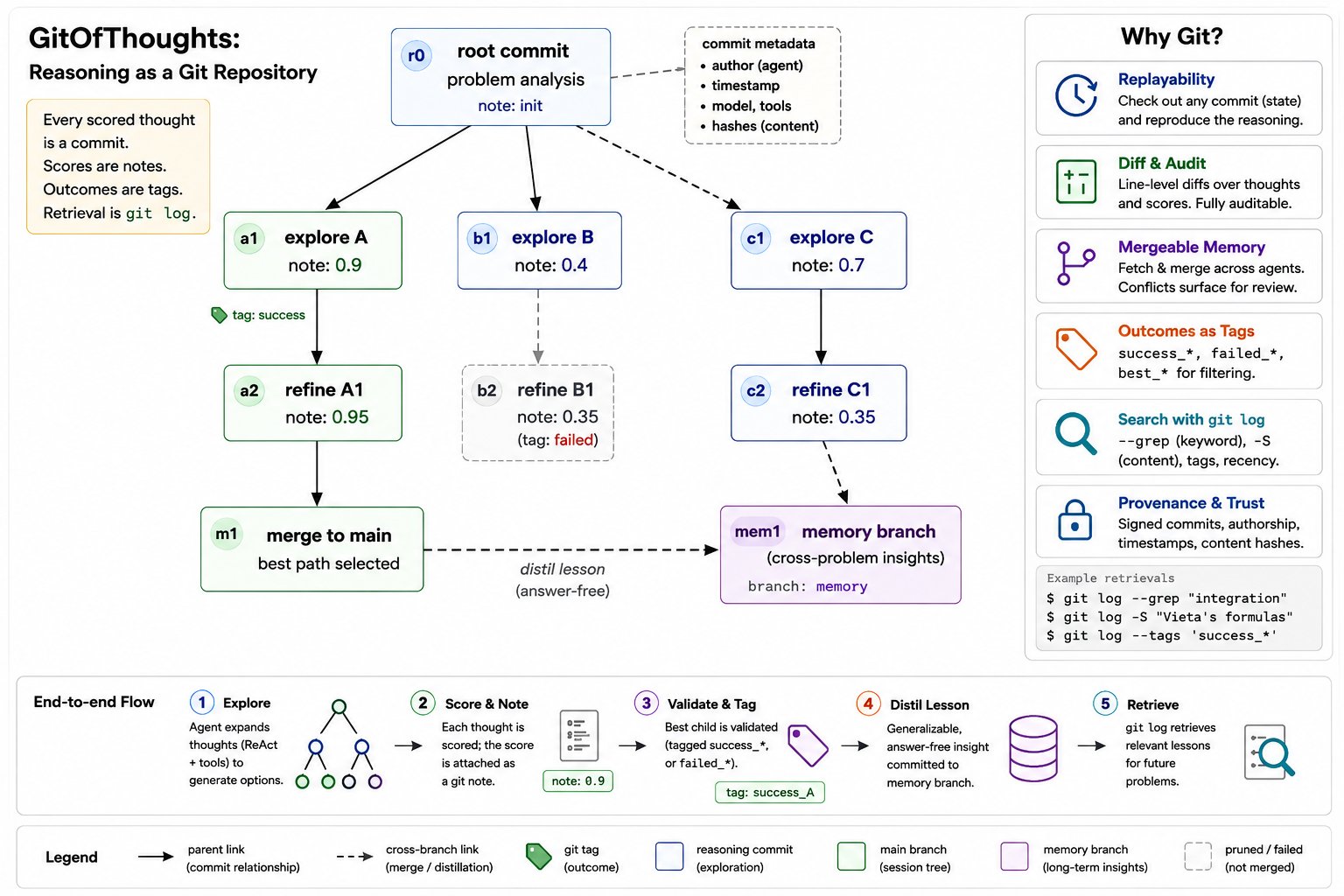}
\caption{The reasoning tree is a git repository. Each scored thought is a commit with author, timestamp, and content-hash metadata; scores are git notes; validation outcomes are tags (\texttt{success\_*}, \texttt{failed\_*}); pruned attempts remain in history rather than vanishing. The winning path merges to \texttt{main}, an answer-free lesson is distilled to a long-lived \texttt{memory} branch, and retrieval is \texttt{git log} (\texttt{-{}-grep}, \texttt{-S}, tag filters) over the agent's own history. Right: the operational properties this buys. Bottom: the end-to-end flow from exploration to retrieval.}
\label{fig:repo}
\end{figure*}

A reasoning tree shares every structural invariant git was designed for: immutable nodes, parent links, content addressing by hash, labelling, branching, and search. Table~\ref{tab:mapping} gives the one-to-one mapping; Figure~\ref{fig:repo} shows it in action. Every property git provides (immutable history, distributed replication, cryptographic verifiability, three-way merge, content-defined deduplication, decades of tooling) becomes a property of the reasoning trace at near-zero engineering cost.

\begin{table}[t]
\centering\small
\caption{Reasoning concept $\leftrightarrow$ git primitive.}
\label{tab:mapping}
\begin{tabular}{@{}ll@{}}
\toprule
\textbf{Reasoning concept} & \textbf{git primitive} \\
\midrule
Scored reasoning node & commit \\
Node identity & SHA-1 commit hash \\
Refinement / parent & commit parent edge \\
Combined score & git notes \\
Validation outcome & git tag (\texttt{success\_*}, \texttt{failed\_*}) \\
Exploration path & branch \\
Session vs.\ cross-session & \texttt{master} vs.\ \texttt{memory} \\
Keyword retrieval & \texttt{git log -{}-grep} \\
Content retrieval & \texttt{git log -S} (pickaxe) \\
Cross-agent merge & \texttt{git fetch} \& \texttt{git merge} \\
Reproducible artefact & \texttt{git bundle} \\
Signed trace & \texttt{git commit -S} \\
\bottomrule
\end{tabular}
\end{table}

\subsection{Why not a database, or a JSONL log with grep?}
\label{sec:whynot}

A natural objection is that SQLite, a vector store, or, the strongest modern strawman, \emph{an append-only JSONL log plus \texttt{ripgrep} and SHA-256 hashing} could play the same role, faster. The JSONL strawman does replicate much of git cheaply: append-only history, content hashes, and lexical retrieval. The residual that git uniquely supplies is the operational layer: (i) a tested \emph{three-way merge} with conflict surfacing, the one structurally unique operation for combining two agents' memories; (ii) signed authorship (\texttt{commit -S}) and native refs/tags; (iii) bit-identical reproduction via content addressing and \texttt{git bundle}; (iv) free pack-file deduplication; and (v) an unmatched tooling ecosystem (blame, bisect, hosting, CI) that a bespoke log must rebuild. These are engineering trade-offs, not accuracy claims, and we don't conflate the two. The empirical question (holding the agent fixed and swapping only the substrate, does git retrieve priors as usefully as a semantic index, and at what cost?) is answered head-on in \S\ref{sec:null}: no substrate buys accuracy on novel problems, so the operational properties become the deciding factor. The merge operation, git's strongest structural differentiator, now has a functional demonstration: a cross-process merge through a central bare repository ran end-to-end, and injected contradictory lessons surfaced as merge conflicts that simple concatenation silently retained (\S\ref{sec:headline}). Whether merged memory also helps accuracy is a pre-registered experiment still to run (\S\ref{sec:future}).

\subsection{Reasoning pipeline}
\label{sec:pipeline}

The search machinery that produces the nodes is deliberately conventional. \textbf{Outer loop:} a depth-1, branching-factor-4 tree of thoughts; for multiple-choice questions the four root children are the four candidate answers (A/B/C/D) and the model argues for each. This \emph{MCQ-aware expansion} turned out to be the largest single accuracy lever in the system. \textbf{Inner loop:} each node runs a ReAct loop (max 3 steps) with a calculator, a \texttt{sympy} solver, and a \texttt{pulp} LP solver; web search is disabled in benchmark mode. \textbf{Scoring:} $s = 0.6\,s_{\text{local}} + 0.4\,s_{\text{cross}}$; the best child is validated, tagged \texttt{success/best} or \texttt{failed}, and at most one re-expansion is allowed with typed-failure context. The score-and-tag step is where GitOfThoughts deviates from vanilla tree-of-thoughts.

\subsection{Git-native memory}
Each problem is solved in its own ephemeral repository. A commit-on-score hook writes four files per node (\texttt{thought.md}, \texttt{scores.json}, \texttt{trace.jsonl}, \texttt{metadata.json}), commits with a structured message, attaches a git note with the score, and tags the outcome. The \texttt{master} branch accumulates the session tree; a long-lived \texttt{memory} branch holds cross-problem insights. Retrieval uses only stock git: keyword (\texttt{-{}-grep}), content (\texttt{-S} pickaxe), outcome filter (\texttt{git tag -l 'success\_*'}), frontier (\texttt{git log master..memory}), ranked by a confidence-weighted score $\rho = \alpha\,s_{\text{comb}} + \beta\,\text{tag} + \gamma\,\text{recency}$ with $(\alpha,\beta,\gamma)=(0.7,0.2,0.1)$.

\subsection{Auditability properties}
\label{sec:audit}

These come with the substrate rather than from extra engineering. \textbf{Replayability:} any SHA reconstructs a full reasoning state via \texttt{git checkout}. \textbf{Incident review:} \texttt{git diff success\_X failed\_Y -- thought.md} gives a line-level diff over the actual reasoning text. \textbf{Mergeable memory:} \texttt{git fetch peer \&\& git merge peer/memory} synchronizes two agents' experience, surfacing conflicts for adjudication. \textbf{Searchable archives:} 10{,}000 sessions are 10{,}000 repos, queryable with one shell loop. \textbf{Fairness via \texttt{git log -S}:} we audit our own GPQA run with the framework's own primitive (Table~\ref{tab:audit}), the same primitive an external reviewer would use. Pre-registrations in this paper are themselves git commits: the stronger-model replication's decision rules were committed before the 32B model served its first token, and the commit history is the audit trail.

\begin{table}[t]
\centering\small
\caption{Fairness audit (GPQA-Diamond). Each row is one git invocation.}
\label{tab:audit}
\begin{tabular}{@{}lc@{}}
\toprule
\textbf{Check} & \textbf{Result} \\
\midrule
Same gold letters across 3 arms? & 100/100 \\
\texttt{expected\_answer} in any commit? & 0 hits \\
Verbatim gold letter in any trace? & 0/30 \\
Fresh memory per question? & verified \\
Random-seed determinism (re-run)? & bit-identical \\
\bottomrule
\end{tabular}
\end{table}

\section{The Memory Question: a Robust Null}
\label{sec:null}

Does cross-problem memory, in any substrate, improve accuracy on novel problems? We answer with three escalating experiments, all behind one interface, with pre-registered hypotheses. A note on statistical power: the $n{=}40$ results (\S\ref{sec:fiveway}--\ref{sec:inside}) are exploratory, with $\pm{\sim}12\pp$ confidence intervals that can only detect large effects, and exist to generate hypotheses that the high-powered $n{=}500$ study (\S\ref{sec:math500}) and the pre-registered replications then confirm or kill. We keep the small-$n$ trends in the paper because one of them produced the central methodological lesson of the whole study.

\subsection{A pluggable \texttt{MemoryBackend}}
Every memory read/write in the agent flows through one object with safe no-op defaults (\texttt{add\_insight/failure/ success}, \texttt{get\_relevant}, \texttt{get\_summary}, cost meters); the chosen backend is injected with a one-line change and tool code is untouched. Five implementations:
\begin{itemize}\setlength\itemsep{1pt}
\item \textbf{none}: no cross-problem memory (ablation control);
\item \textbf{markdown}: a human-readable \texttt{memory.md} with deduplication and confidence weighting;
\item \textbf{git}: one shared repo; insights are commits on a \texttt{memory} branch; retrieval is \texttt{git grep};
\item \textbf{vector}: all-MiniLM-L6-v2 embeddings \citep{reimers2019sbert} in a Chroma index, cosine nearest-neighbour;
\item \textbf{graph}: a \texttt{networkx} lesson/concept graph; spread-activation retrieval (associative, not nearest-neighbour).
\end{itemize}

\subsection{Five substrates, controlled transfer (exploratory, $n{=}40$)}
\label{sec:fiveway}

\textbf{Protocol.} To isolate retrieval from write-path noise, all backends ingest \emph{identical} knowledge. Per benchmark: (i) \emph{study}: solve $K$ study problems, distilling one generalizable, answer-free lesson each into a shared \texttt{lessons.jsonl}; (ii) \emph{ingest}: every backend replays the same lessons; (iii) \emph{test}: solve $M$ held-out problems with each backend injected read-only. Splits are deterministic, domain-stratified (study and test share domains, enabling transfer), and disjoint (audited: zero overlap). The primary agent is a single-shot retrieval-augmented solver; benchmarks are GPQA-Diamond and MATH-500; backbone Qwen3.5-9B served via vLLM \citep{kwon2023vllm} on one NVIDIA L40S.

\textbf{Cost (measured).} The substrates differ sharply in cost but all are cheap in absolute terms (Table~\ref{tab:cost}): git pays $\sim$15\,ms per write (a commit) and $\sim$48\,ms per read (grep over history), the same order as the embedding index's read (20\,ms).

\begin{table}[t]
\centering\small
\caption{Per-backend cost (GPQA, 40 ingested lessons).}
\label{tab:cost}
\begin{tabular}{@{}lrrr@{}}
\toprule
\textbf{Backend} & \textbf{Write (ms)} & \textbf{Read (ms)} & \textbf{Storage} \\
\midrule
none & --- & --- & 0 \\
markdown & 0.51 & 0.00 & 69\,KB \\
graph & 0.14 & 0.35 & 184\,KB \\
git & 15.4 & 48.0 & 191\,KB \\
vector & 30.4 & 20.3 & 656\,KB \\
\bottomrule
\end{tabular}
\end{table}

\textbf{Pre-registered hypotheses.} (H1) accumulated memory beats none, more on MATH-500 (recurring techniques) than GPQA; (H2) semantic backends beat lexical ones; (H3) git's value is the engineering trade-off at accuracy parity, not top raw retrieval.

\textbf{Findings.} The data discipline all three (Table~\ref{tab:exp2}). H1 is not supported: on GPQA every CI includes zero; on MATH-500 every backend lands 2.5--5$\pp$ below the no-memory control. H2 has only a weak, local signal: vector is the single positive cell ($+10\pp$, CI $\ci{-2.5}{+22.5}$ does not clear zero at $n{=}40$), a trend rather than a result. H3 is supported: git is statistically indistinguishable from the best backend on both benchmarks while being the only substrate that supplies the auditability properties of \S\ref{sec:audit}.

\begin{table}[t]
\centering\small
\caption{Transfer gain $\Delta$ vs.\ none (single-shot agent, $n{=}40$/cell, paired bootstrap 95\% CI). Baselines: GPQA none $=52.5\%$, MATH-500 none $=57.5\%$. Exploratory scale.}
\label{tab:exp2}
\begin{tabular}{@{}llrr@{}}
\toprule
\textbf{Bench.} & \textbf{Backend} & \textbf{Acc.} & $\Delta\pp$ [95\% CI] \\
\midrule
\multirow{4}{*}{GPQA}
 & markdown & 52.5\% & $+0.0$ \ci{-12.5}{+12.5} \\
 & git & 50.0\% & $-2.5$ \ci{-15.0}{+10.0} \\
 & vector & 62.5\% & $+10.0$ \ci{-2.5}{+22.5} \\
 & graph & 50.0\% & $-2.5$ \ci{-12.5}{+7.5} \\
\midrule
\multirow{4}{*}{MATH-500}
 & markdown & 55.0\% & $-2.5$ \ci{-10.0}{+5.0} \\
 & git & 52.5\% & $-5.0$ \ci{-12.5}{+0.0} \\
 & vector & 52.5\% & $-5.0$ \ci{-12.5}{+0.0} \\
 & graph & 52.5\% & $-5.0$ \ci{-12.5}{+0.0} \\
\bottomrule
\end{tabular}
\end{table}

\subsection{Memory inside the agent, and a trend that died}
\label{sec:inside}

We next wire the backends inside the full agent (lessons retrieved into the node prompt; distilled, answer-free lessons written after each problem) and test two regimes: cross-problem (GPQA, read-only frozen study memory from 30 disjoint problems) and cross-episode (ScienceWorld, $n{=}66$/arm, memory accumulating sequentially so the learning curve is the signal).

At $n{=}40$, git showed the largest gain of any backend ($+15\pp$), but the confidence interval included zero, so we treated it as a trend and ran a confirmation at $n{\approx}100$ before reporting it. The trend did not replicate (Table~\ref{tab:exp3}): at $n{=}98$, git collapses to $+1.0\pp$ $\ci{-10.2}{+11.2}$ and the backend ranking reshuffles (markdown worst$\to$best, vector best$\to$worst, git best$\to{\approx}$none). All CIs straddle zero. The instability of the ranking across samples is itself strong evidence that no backend reliably helps at this scale and model; the $n{=}40$ $+15$ reflected small-sample luck, exactly the failure mode the pre-registered larger-$n$ check exists to catch.

ScienceWorld is a separate case. No backend beats no-memory ($\Delta\le 1.5\pp$, all CIs cross zero) and no learning curve emerges, but the scaffolded 9B sits at ${\sim}12\%$ absolute, versus SwiftSage's 84.7 \citep{lin2023swiftsage}. That is a floor, not a fair test. A null from a floored agent says ``this agent could not demonstrate a benefit,'' not ``memory does not help cross-episode.'' We therefore exclude ScienceWorld from the headline null claim and report it as an inconclusive arm pending a stronger base agent.

\begin{table}[t]
\centering\small
\caption{Memory inside the agent, $\Delta$ vs.\ none (paired bootstrap 95\% CI). GPQA $=$ accuracy; ScienceWorld $=$ mean task score (floored; see text).}
\label{tab:exp3}
\begin{tabular}{@{}llrr@{}}
\toprule
\textbf{Setting} & \textbf{Backend} & \textbf{Score} & $\Delta$ [95\% CI] \\
\midrule
\multirow{4}{*}{\shortstack[l]{GPQA\\($n{=}40$, expl.)}}
 & none & 57.5\% & baseline \\
 & markdown & 50.0\% & $-7.5$ \ci{-25}{+10} \\
 & vector & 62.5\% & $+5.0$ \ci{-7.5}{+17.5} \\
 & git & 72.5\% & $+15.0$ \ci{-2.5}{+32.5} \\
\midrule
\multirow{4}{*}{\shortstack[l]{GPQA\\($n{=}98$, conf.)}}
 & none & 57.1\% & baseline \\
 & markdown & 62.2\% & $+5.1$ \ci{-5.1}{+15.3} \\
 & vector & 48.0\% & $-9.2$ \ci{-20.4}{+2.0} \\
 & git & 58.2\% & $+1.0$ \ci{-10.2}{+11.2} \\
\midrule
\multirow{4}{*}{\shortstack[l]{ScienceWorld\\($n{=}66$, floored)}}
 & none & 12.3\% & baseline \\
 & markdown & 12.1\% & $-0.2$ \ci{-3.8}{+3.0} \\
 & vector & 13.2\% & $+0.9$ \ci{-1.7}{+3.7} \\
 & git & 13.9\% & $+1.5$ \ci{-0.5}{+4.0} \\
\bottomrule
\end{tabular}
\end{table}

\subsection{Scaling the question: MATH-500 at $n{=}500$, with mechanism ablations}
\label{sec:math500}

We now test at the largest scale, on a second backbone (Qwen2.5-7B-Instruct \citep{yang2024qwen25}), on the benchmark most favourable to memory: MATH-500, whose within-subject problems reuse transferable methods (Vieta's formulas, casework, telescoping), unlike GPQA's disjoint facts. Memory corpus: 2{,}000 disjoint MATH-train problems with gold worked solutions (leakage-audited against the test set); retrieval top-3; a \texttt{sympy}-hardened answer checker applied identically to every arm. Three added arms turn the substrate comparison into a mechanism test: \textbf{static few-shot} (3 fixed exemplars; controls ``having examples'' vs.\ ``retrieving relevant ones''), \textbf{self-consistency} (5$\times$ majority vote; the standard test-time lever), and \textbf{content/relevance ablations} on the best retriever (full worked solution vs.\ answer-only vs.\ answer-free lesson; subject-filtered retrieval).

\begin{table}[t]
\centering\small
\caption{MATH-500, $n{=}500$, Qwen2.5-7B-Instruct. $\Delta$ vs.\ none (66.6\%), paired bootstrap 95\% CI; \sig{} $=$ CI excludes 0.}
\label{tab:math500}
\begin{tabular}{@{}lrr@{}}
\toprule
\textbf{Arm} & \textbf{Acc.} & $\Delta$ vs.\ none [95\% CI] \\
\midrule
none (zero-shot CoT) & 66.6\% & baseline \\
\textbf{sc5 (self-consistency)} & \textbf{70.0\%} & $\mathbf{+3.4}$ \ci{+0.6}{+6.2}\,\sig \\
static few-shot (3 fixed) & 67.0\% & $+0.4$ \ci{-2.4}{+3.4} \\
markdown (lexical) & 66.8\% & $+0.2$ \ci{-2.6}{+3.2} \\
vector (semantic) & 68.2\% & $+1.6$ \ci{-1.4}{+4.6} \\
git (grep) & 65.8\% & $-0.8$ \ci{-4.2}{+2.4} \\
vector, subject-filtered & 66.2\% & $-0.4$ \ci{-3.6}{+2.8} \\
vector, answer-only & 67.6\% & $+1.0$ \ci{-2.2}{+4.2} \\
vector, answer-free lesson & 66.2\% & $-0.4$ \ci{-3.4}{+2.6} \\
\bottomrule
\end{tabular}
\end{table}

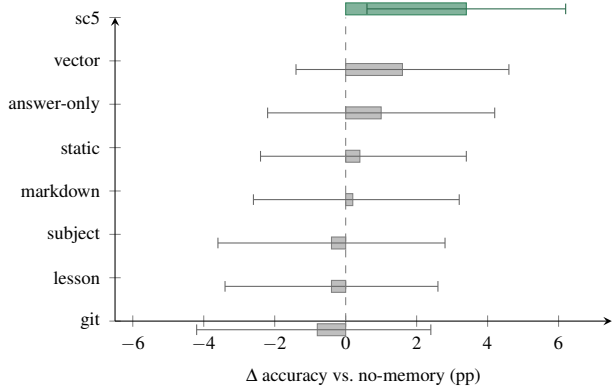
\begin{figure}[t]
\centering
\begin{tikzpicture}
\begin{axis}[
  width=\columnwidth, height=5.6cm,
  xbar, xmin=-6.5, xmax=7.5,
  xlabel={$\Delta$ accuracy vs.\ no-memory (pp)},
  symbolic y coords={git,lesson,subject,markdown,static,answer-only,vector,sc5},
  ytick={git,lesson,subject,markdown,static,answer-only,vector,sc5}, y dir=normal,
  tick label style={font=\scriptsize}, label style={font=\scriptsize},
  axis lines=left, clip=false,
  every axis plot/.append style={bar width=4.5pt},
]
\addplot+[xbar, fill=black!25, draw=black!50, error bars/.cd, x dir=both, x explicit,
  error bar style={black!60}] coordinates {
  (-0.8,git)  += (3.2,0) -= (3.4,0)
  (-0.4,lesson) += (3.0,0) -= (3.0,0)
  (-0.4,subject) += (3.2,0) -= (3.2,0)
  (0.2,markdown) += (3.0,0) -= (2.8,0)
  (0.4,static) += (3.0,0) -= (2.8,0)
  (1.0,answer-only) += (3.2,0) -= (3.2,0)
  (1.6,vector) += (3.0,0) -= (3.0,0)
};
\addplot+[xbar, fill=gitgreen!60, draw=gitgreen, error bars/.cd, x dir=both, x explicit,
  error bar style={gitgreen!80!black}] coordinates {
  (3.4,sc5) += (2.8,0) -= (2.8,0)
};
\draw[dashed, black!50] (axis cs:0,git) -- (axis cs:0,sc5);
\end{axis}
\end{tikzpicture}
\caption{What actually moves accuracy (MATH-500, $n{=}500$). Per-arm $\Delta$ vs.\ no-memory with paired bootstrap 95\% CIs. Only self-consistency (green) clears zero; every memory substrate and the static-few-shot control sit within noise.}
\label{fig:math500}
\end{figure}

\textbf{Findings.} Self-consistency is the only arm that significantly beats no-memory ($+3.4\pp$\sig); no retrieval substrate does (Table~\ref{tab:math500}, Fig.~\ref{fig:math500}). Four results pin down why, one of which led us to retract an earlier subgroup hypothesis:
\begin{enumerate}\setlength\itemsep{1pt}
\item \textbf{Relevance does not drive a gain.} Forcing same-subject retrieval ($-0.4$) did not beat unconstrained vector retrieval ($+1.6$). An earlier subgroup signal (``problems whose retrieved exemplar shares the subject score higher'') turned out to be confounded: semantically similar exemplars happen to share a subject, and forcing the subject label discards the better global match. We had hypothesized relevance was the lever; the controlled arm refutes it.
\item \textbf{Content barely matters.} Full worked solution ($+1.6$) $\approx$ answer-only ($+1.0$) $>$ distilled lesson ($-0.4$), all within noise; the model is not extracting transferable method from the exemplars.
\item \textbf{Having examples is not the lever.} Static few-shot ($+0.4$) $\approx$ none.
\item \textbf{No online learning curve.} With a git memory accumulating the agent's own verified solutions across the 500-problem stream, accuracy by quartile runs $70.4 \to 60.0 \to 67.2 \to 67.2\%$, slightly down, even as subject-relevant retrievals grow.
\end{enumerate}

\textbf{Cross-model robustness.} Re-running the GPQA memory comparison on this second backbone ($n{=}98$) reproduces the null and the pattern: no substrate beats no-memory (none 40.8\%; markdown $-4.1$, git $-2.0$, vector $-8.2$; all CIs cross zero), with vector worst on both models and git ${\approx}$ none on both. The null therefore holds across two backbones (Qwen3.5-9B, Qwen2.5-7B-Instruct), two benchmarks, and up to $n{=}500$. We state the scope of this claim precisely: the two full-suite backbones are adjacent in size and from different model families, and the 32B (\S\ref{sec:copyability}) runs only the method-transfer arm; this is a robustness check, not a scaling law.

\section{Where the Null Breaks: the Copyability Threshold}
\label{sec:copyability}

Every test so far lived in the disjoint-problem regime. But the systems that plausibly benefit from memory (support tickets, recurring bugs, repeated workflows) retrieve near-duplicate past cases. We therefore vary the test$\leftrightarrow$memory similarity directly: for 200 hard (Level 4--5) MATH seeds we inject a single worked example at four similarity tiers and measure $\Delta$ vs.\ no-memory (7B baseline 53.0\%; cosine via MiniLM).

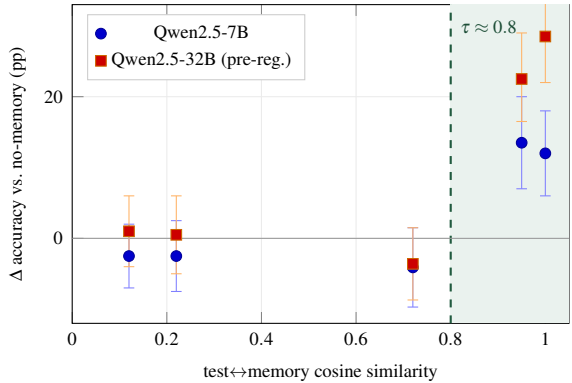
\begin{figure}[t]
\centering
\begin{tikzpicture}
\begin{axis}[
  width=\columnwidth, height=5.8cm,
  xlabel={test$\leftrightarrow$memory cosine similarity},
  ylabel={$\Delta$ accuracy vs.\ no-memory (pp)},
  xmin=0, xmax=1.05, ymin=-12, ymax=33,
  tick label style={font=\scriptsize}, label style={font=\scriptsize},
  legend style={font=\scriptsize, at={(0.03,0.97)}, anchor=north west, draw=black!20},
  grid=major, grid style={black!8},
]
\fill[gitgreen!10] (axis cs:0.8,-12) rectangle (axis cs:1.05,33);
\draw[dashed, gitgreen!70!black, thick] (axis cs:0.8,-12) -- (axis cs:0.8,33)
  node[pos=0.93, right, font=\scriptsize, text=gitgreen!60!black] {$\tau\approx0.8$};
\draw[black!40] (axis cs:0,0) -- (axis cs:1.05,0);
\addplot+[only marks, mark=*, mark size=2pt, color=blue!60!black,
  error bars/.cd, y dir=both, y explicit, error bar style={blue!50}]
coordinates {
  (0.12,-2.5) += (0,4.5) -= (0,4.5)
  (0.22,-2.5) += (0,5.0) -= (0,5.0)
  (0.72,-4.1) += (0,5.6) -= (0,5.6)
  (0.95,13.5) += (0,6.5) -= (0,6.5)
  (1.00,12.0) += (0,6.0) -= (0,6.0)
};
\addlegendentry{Qwen2.5-7B}
\addplot+[only marks, mark=square*, mark size=2pt, color=orange!80!black,
  error bars/.cd, y dir=both, y explicit, error bar style={orange!60}]
coordinates {
  (0.12,1.0) += (0,5.0) -= (0,5.0)
  (0.22,0.5) += (0,5.5) -= (0,5.5)
  (0.72,-3.6) += (0,5.1) -= (0,5.1)
  (0.95,22.5) += (0,6.5) -= (0,6.0)
  (1.00,28.5) += (0,6.5) -= (0,6.5)
};
\addlegendentry{Qwen2.5-32B (pre-reg.)}
\end{axis}
\end{tikzpicture}
\caption{The copyability threshold. Memory helps only above cosine ${\approx}0.8$ (shaded), where the retrieved case is a near-duplicate. The method band (cosine 0.72: same method, different numbers, non-copyable answer) is null at both scales: a $4.5\times$ larger model steepens the step (doubling the near-duplicate payoff) without unlocking method transfer.}
\label{fig:threshold}
\end{figure}

This is the first positive memory effect in this paper, and the boundary is clean (Table~\ref{tab:sweep}, Fig.~\ref{fig:threshold}). A retrieved near-duplicate or paraphrase (cosine $\gtrsim 0.85$) gives $+12$ to $+13.5\pp$ with CIs clearing zero, while a same-subject or unrelated example (cosine $\le 0.22$) gives nothing. Pooling by cosine, the gain clears zero only in the top bin ($\ge 0.85$: $+12.7$ $\ci{+8.4}{+17.3}$\sig). We identify this as a copyability threshold $\tau\approx 0.8$: above it, memory reliably improves accuracy; below it, effects are indistinguishable from noise. This explains every prior null (GPQA/MATH cross-problem exemplars sit at cosine ${\sim}0.1$--$0.5$, below $\tau$) and bounds when memory pays off: recurring or near-duplicate workloads, not novel problems.

The gain reflects copyability, not method transfer. A dedicated method-transfer arm, in which the memory is a real same-method, different-numbers problem (cosine 0.72; a different answer, hence not copyable; 187/196 memory answers differ from the test's), is null: $-4.1$ $\ci{-9.7}{+1.5}$. We had earlier hypothesized that a worked example would help even when the answer itself differed, since the underlying method was shared; this arm refutes that. The model benefits from near-verbatim recurrence, where the answer is essentially retrievable, not from abstracting a method out of a worked example. Even with the exact solution present, the high-similarity tiers reach only 65\%: the 7B does not blindly copy a retrieved answer either.

\textbf{Does a stronger model transfer method? (pre-registered).} A natural objection is that a 7B is simply too weak to exploit a retrieved method. We re-ran the method-transfer protocol unchanged on Qwen2.5-32B-Instruct (AWQ int4; the largest backbone our 46\,GB GPU serves), fixing the decision rule before the run: method transfer is present iff the paired bootstrap 95\% CI of $\Delta(\text{method}-\text{none})$ clears zero; the test is valid only if the recurrence control stays positive and the baseline is below ceiling. Both validity conditions hold (none $=60.7\%$ vs.\ the 7B's 54.6\% on the same seeds; 190/196 memory answers non-copyable; 0 empty predictions; median 20.9\,s/solve). The null replicates almost exactly, with $\Delta = -3.6$ $\ci{-8.7}{+1.5}$ (core band $[0.60,0.85]$: $-2.4$ $\ci{-8.4}{+3.6}$), vs.\ the 7B's $-4.1$. The near-duplicate control does more than just pass: it amplifies. The 32B exploits near-verbatim memory far better than the 7B (identical $+28.5$ $\ci{+22.0}{+35.0}$\sig, paraphrase $+22.5$ $\ci{+16.5}{+29.0}$\sig, reaching 86\% where the 7B plateaued at 65\%), while same-subject and unrelated tiers stay null. Scale steepens the copyability step without turning retrieved worked examples into transferable method. (AWQ int4 can shift the absolute level, not the paired within-model $\Delta$s; the remaining open variable is a frontier-class model.)

\begin{table}[t]
\centering\small
\setlength{\tabcolsep}{3pt}
\caption{Similarity sweep and method band across scale (MATH, $n{=}200$ hard seeds; method band $n{=}196$ paired). $\Delta$ vs.\ no-memory per tier, paired bootstrap 95\% CI; \sig{} $=$ CI excludes 0. Baselines (none): 7B 53.0\% (sweep) / 54.6\% (method run); 32B 57.5\% / 60.7\%.}
\label{tab:sweep}
\begin{tabular}{@{}lrr@{}}
\toprule
\textbf{Tier (mean cos.)} & $\Delta$ \textbf{7B} [CI] & $\Delta$ \textbf{32B} [CI] \\
\midrule
identical (1.00) & $+12.0$ \ci{+6.0}{+18.0}\,\sig & $+28.5$ \ci{+22.0}{+35.0}\,\sig \\
paraphrase (0.95) & $+13.5$ \ci{+7.0}{+20.0}\,\sig & $+22.5$ \ci{+16.5}{+29.0}\,\sig \\
method band (0.72) & $-4.1$ \ci{-9.7}{+1.5} & $-3.6$ \ci{-8.7}{+1.5} \\
same-subject (0.22) & $-2.5$ \ci{-7.5}{+2.5} & $+0.5$ \ci{-5.0}{+6.0} \\
unrelated (0.12) & $-2.5$ \ci{-7.0}{+2.0} & $+1.0$ \ci{-4.0}{+6.0} \\
\bottomrule
\end{tabular}
\end{table}

\section{What Actually Moves Accuracy}
\label{sec:levers}

\subsection{Test-time sampling}
Self-consistency \citep{wang2023sc} is the one arm in the entire study whose CI clears zero on novel problems ($+3.4\pp$ $\ci{+0.6}{+6.2}$ at $n{=}500$; Table~\ref{tab:math500}). That self-consistency works is not new \citep{snell2024testtime}; the value here is the contrast: under matched conditions, the standard test-time lever clears zero while every memory substrate sits in noise.

\subsection{Test-time architectures: none beat greedy}
Across the memory experiments the selector/aggregator kept emerging as the bottleneck, so we iterated test-time architectures directly on fixed GPQA sets, paired against greedy chain-of-thought: self-consistency (several temperatures), argue-each-option, single-revision, and verifier best-of-$N$. One methodological point matters before the results: at $n{=}20$ the noise floor is $\pm 10\pp$ (greedy itself swung 55\% $\leftrightarrow$ 65\% across identical runs), so single-cycle deltas are noise and we don't crown a winner from them. With that constraint applied, no test-time architecture reliably beats greedy single CoT for this model on GPQA, and self-consistency and revision can hurt (sampling and over-editing break already-correct answers). The one stable qualitative signal is that verifier-based selection is conservative, in that it rarely breaks a correct answer. A pre-registered $n{=}40$ confirmation gives greedy 52.5\% vs.\ verifier best-of-$N$ 55.0\%: a gain of $+2.5\pp$ (one additional correct answer) at $6.4\times$ the token cost, well within noise and not worth the cost.

\subsection{The system headline, with its confound stated}
\label{sec:headline}

For completeness we report the original system result, with the caveat it needs. The middle arm is ReTreVal \citep{retreval2026}, a reasoning-tree framework with validation, critique scoring, and cross-problem memory; GitOfThoughts re-instruments ReTreVal-style tree search on the git substrate of \S\ref{sec:design}, so this comparison is against a full tree-search system, not plain sampling. All arms use identical model and data; fresh-memory mode on all memory arms conservatively denies GitOfThoughts any cross-problem advantage. Results on GPQA-Diamond (Qwen3.5-9B, 100 questions):

\begin{table}[h]
\centering\small
\caption{GPQA-Diamond system comparison. \textbf{Caution: wall-clock budgets differ across arms (60\,s / 180\,s / 600\,s), so this comparison confounds method with compute.} A compute-matched control is future work (\S\ref{sec:limits}).}
\label{tab:headline}
\begin{tabular}{@{}lrrr@{}}
\toprule
\textbf{Arm} & \textbf{Acc.} & \textbf{Budget} & \textbf{Per-Q} \\
\midrule
Vanilla (single call) & 33.0\% & 60\,s & 0.7\,s \\
ReTreVal (5$\times$ SC) \citep{retreval2026} & 34.0\% & 180\,s & 127.9\,s \\
GitOfThoughts & 47.0\% & 600\,s & 469.8\,s \\
\bottomrule
\end{tabular}
\end{table}

\begin{table*}[t]
\centering\small
\caption{Every experiment, its setup, and its verdict.}
\label{tab:glance}
\begin{tabular}{@{}p{4.2cm}p{5.2cm}p{6.8cm}@{}}
\toprule
\textbf{Experiment} & \textbf{What we tested} & \textbf{Verdict} \\
\midrule
1. GPQA headline (9B) & GitOfThoughts system vs.\ vanilla / ReTreVal (5$\times$ SC) & 47\% vs.\ 33\% ($+14\pp$), but budgets differ across arms and the gain is driven by MCQ-aware expansion $+$ compute, not git or memory. \\
2. Five-substrate study$\to$test (9B, $n{=}40$, exploratory) & none / markdown / vector / graph / git & No substrate beats none; git at accuracy parity with the best while uniquely auditable. \\
3. Memory inside the agent (GPQA $n{=}40{\to}98$; ScienceWorld) & cross-problem $+$ cross-episode transfer & A $+15\pp$ git trend did not replicate ($\to +1.0$, ranking reshuffled). ScienceWorld is floored ($\sim$12\%) and excluded from the headline null. \\
4. Test-time architectures (9B) & SC / argue / revise / verifier best-of-$N$ & None beat greedy CoT; SC and revision can hurt. The selector is the bottleneck. \\
5. MATH-500 $n{=}500$ $+$ ablations (7B) & 9 arms: substrates, static few-shot, SC, content, subject filter, online & Self-consistency $+3.4\pp$\sig{} is the only significant lever; relevance and content refuted; no online learning curve. \\
Cross-model robustness (7B, GPQA $n{=}98$) & re-run the memory null on a 2nd backbone & Null holds across two backbones, two benchmarks, and up to $n{=}500$ (scope stated precisely in \S\ref{sec:math500}). \\
Recurrence sweep (7B, MATH, $n{=}200$) & vary test$\leftrightarrow$memory similarity & Regime-dependent: near-duplicate retrieval (cos $\gtrsim 0.8$) gives $+12$ to $+13.5\pp$\sig; same-subject/disjoint gives nothing ($\tau\approx0.8$). \\
Method transfer (7B, MATH, $n{=}200$) & same method, different numbers (not copyable) & Null ($-4.1$ \ci{-9.7}{+1.5}): the benefit is answer retrieval, not method transfer. \\
Method transfer at scale (32B, $n{=}196$, pre-registered) & rerun on a $4.5\times$ backbone & Null replicates ($-3.6$ \ci{-8.7}{+1.5}) while the near-duplicate control more than doubles ($+22.5$ to $+28.5\pp$\sig); scale steepens copyability, not transfer. \\
Durable execution (Hatchet) & each node as a durable cloud task & Results identical by construction; cost ${\sim}249$\,ms/node dispatch: durability, not accuracy. Functional distributed merge run: 5/5 injected contradictions surface as conflicts; concatenation stays silent. Accuracy phase pending. \\
\bottomrule
\end{tabular}
\end{table*}

Our own ablations show that the $+14\pp$ gain decomposes into MCQ-aware expansion (the largest single lever) plus the larger compute budget, not git and not memory. Of the 53 misses, 46 are timeouts, concentrated in Chemistry's multi-step synthesis chains.\footnote{On the 54 questions that completed within budget, accuracy is 87.0\% (47/54). We report this only as a descriptive footnote: conditioning on completion selects the easier-to-finish problems (timeouts correlate with problem length and difficulty), so this number is upward-biased and is not an estimate of accuracy under a larger budget. The unbiased version of this claim requires re-running with a higher cap, which we list as future work.} For a fixed model already near its competence ceiling, test-time compute is not the lever; base-model strength is.

\subsection{Durable execution: measured overhead, not accuracy}
To verify the substrate is production-real and not merely a prototype, each reasoning node can also run as a durable task on a hosted Hatchet queue \citep{hatchet2024} (the executor seam of \S\ref{sec:pipeline} swapped for a cloud backend; git remains the shared state). On a mock harness ($n{=}12$ runs, $K{=}4$ branches), node counts and results are identical by construction; the only added cost is ${\sim}249$\,ms/node of dispatch latency, which amortizes once each node's real work (an LLM call, on the order of seconds) dominates. Orchestration choices, like memory substrate choices, do not move accuracy; they buy durability and crash-resume guarantees.

The distributed variant, which realizes the mergeable-memory claim across processes, has now been exercised end-to-end in a functional run: two worker processes cloned a central bare repository, accumulated lessons on separate branches over disjoint domains, and the orchestrator merged them, with zero conflicts, as expected for disjoint content. Five deliberately contradictory lessons were then injected; all five surfaced as genuine git merge conflicts, while a concatenation control silently retained all five contradictions with no error signal. One caveat surfaced along the way: with content-hashed lesson filenames, merges are conflict-free by construction and contradictions can coexist silently, the same failure mode as concatenation; conflict surfacing requires a keyed layout (one file per topic) so the merge encodes ``same slot, different content.'' Git supports both layouts and surfaces the contradiction exactly when the layout encodes it; concatenation cannot, in either layout. This run exercised the plumbing with stub lessons; the accuracy phase with model-distilled lessons is pre-registered and pending (\S\ref{sec:future}).

\section{All Experiments at a Glance}

Table~\ref{tab:glance} summarizes every experiment, including the trend we retracted and the hypothesis we refuted. Two findings hold across all of them: the accuracy levers are test-time sampling and base-model strength, and memory pays only above the copyability threshold, where git's auditability and mergeability come at accuracy parity.

\section{Discussion and Limitations}
\label{sec:limits}

\textbf{Compute trade-off and the headline confound.} GitOfThoughts buys its system gain at large wall-clock cost ($\sim$470\,s/question), and the arms in Table~\ref{tab:headline} run different budgets. The practical case is high-stakes reasoning, not latency-sensitive serving. Until a compute-matched control runs (\S\ref{sec:future}), the 47\% should be read as a system-plus-compute result.

\textbf{What the null does and does not cover.} The null is established for short, distilled, answer-free lessons and worked-example retrieval, on two adjacent open-weight backbones, up to $n{=}500$. It does not cover frontier-class models, rich episodic memory (full reasoning traces rather than one-shot distilled lessons, which every arm here under-tests), or multi-agent settings; \S\ref{sec:future} lays out the experiments that would close each gap.

\textbf{Other open items.} ScienceWorld's floor ($\sim$12\%) and its uniform action-mask scaffolding; a single seed for wall-clock budgets; AWQ int4 quantization on the 32B (affects the absolute level, not paired within-model $\Delta$s); and the controlled-transfer ``identical knowledge'' assumption holds only for the simple ingest path.

\section{Future Work}
\label{sec:future}

Three experiments follow directly from this paper, each to be pre-registered before the first model call. First, and highest priority, merged-memory accuracy: the functional phase is complete (\S\ref{sec:headline}); what remains is the pre-registered accuracy phase, in which two agents accumulate model-distilled memory over disjoint domains, the orchestrator merges, and merged memory is compared against each solo memory and a naive concatenation on a mixed-domain test set. Given the copyability threshold, we expect the accuracy deltas to be small or null; the operational comparison against concatenation is the primary endpoint. Second, a compute-matched system baseline: every GPQA arm at the same wall-clock budget, including budget-saturating self-consistency, which either validates the system headline or converts it into a clean negative result. Third, moving the threshold: frontier-class backbones together with rich episodic memory (full reasoning traces rather than distilled lessons) are the two variables most likely to lower $\tau$, and we plan to vary them independently, since either alone could rescue method transfer. Further out: an unbiased high-budget headline run, distributed durable execution across machines, and cross-episode transfer on a non-floored agent.

\section{Conclusion}

LLM reasoning is the last unversioned software process. GitOfThoughts makes every scored reasoning node a git commit (notes for scores, tags for outcomes, branches for memory, \texttt{git log} as the retrieval API), buying provenance, replay, diff, merge, and reproducibility at near-zero cost. Beyond the substrate, this paper is an exercise in pre-registered evaluation across five experiments, two backbones, and up to $n{=}500$, and its sharpest result is a boundary: memory helps only above a copyability threshold ($\tau\approx0.8$), where the retrieved case is a near-duplicate and the gain ($+12$ to $+13.5\pp$\sig{} at 7B; $+22.5$ to $+28.5\pp$\sig{} at 32B) is answer retrieval, not method transfer. A $4.5\times$ scale-up steepens that step without unlocking abstraction. Below the threshold, no substrate (git, vector, graph, markdown) reliably moves accuracy; the one general lever is test-time sampling ($+3.4\pp$\sig). The case for git-as-substrate is therefore auditability, provenance, diff, merge, and reproducibility at accuracy parity, not a retrieval boost. Two questions remain open: whether a frontier-class model with richer episodic memory lowers $\tau$ enough to enable real method transfer, and whether git-mergeable multi-agent memory, the substrate's one structurally unique operation, delivers gains that single-agent retrieval cannot.

\bibliographystyle{plainnat}
{\small
\bibliography{references}
}

\end{document}